\documentclass[11pt,a4paper]{article}
\usepackage[hyperref]{acl2019}
\usepackage{times}
\usepackage{latexsym}

\usepackage{url}

\usepackage{amsmath}
\usepackage{amssymb}
\usepackage{mathtools}
\usepackage{accents}
\usepackage{enumerate}
\usepackage{multirow}

\usepackage{booktabs}
\usepackage{graphicx}
\usepackage{graphics}
\usepackage[ruled,linesnumbered,boxed,lined]{algorithm2e}
\usepackage{makecell}

\usepackage[utf8]{inputenc}

\definecolor{realpurple}{RGB}{87, 6, 140}

\newcommand{\lrec}{L_{\mathit{rec}}}
\newcommand{\ladv}[1]{L_{\mathit{adv}_{#1}}}
\newcommand{\lcyc}{L_{\mathit{cyc}}}
\newcommand{\lpara}{L_{\mathit{para}}}
\newcommand{\llang}{L_{\mathit{lang}}}
\newcommand{\ladvp}[1]{L_{\mathit{adv}'_{#1}}}
\newcommand{\acc}{\mathrm{Acc}}
\newcommand{\gsim}{\mathrm{Sim}}
\newcommand{\met}{\mathrm{Met}}
\newcommand{\pp}{\mathrm{PP}}
\newcommand{\gm}{\mathrm{GM}}
\newcommand{\hcirc}{\accentset{\circ}{\mathbf{h}}}

\let\norm\undefined 
\DeclarePairedDelimiter\norm{\lVert}{\rVert}

\usepackage{graphicx}
\usepackage{subcaption}
\usepackage{mwe}

\usepackage[bottom]{footmisc}

\aclfinalcopy % Uncomment this line for the final submission

\makeatletter
\renewcommand*{\@fnsymbol}[1]{\ensuremath{\ifcase#1\or \mathsection\or \dagger\or \ddagger\or
    \mathsection\or \mathparagraph\or \|\or **\or \dagger\dagger
    \or \ddagger\ddagger \else\@ctrerr\fi}}
\makeatother

\title{Unsupervised Evaluation Metrics and Learning Criteria\\for Non-Parallel Textual Transfer}

\author{Richard Yuanzhe Pang$^1$\thanks{~\ Work completed while the author was a student at the University of Chicago and a visiting student at Toyota Technological Institute at Chicago.} \ \ \ \ \ \ \ Kevin Gimpel$^2$ \\
$^1$New York University, New York, NY 10011, USA \\
$^2$Toyota Technological Institute at Chicago, Chicago, IL 60637, USA \\
{\tt yzpang@nyu.edu, kgimpel@ttic.edu}}

\date{}

\begin{document}
\maketitle
\begin{abstract}
We consider the problem of automatically generating textual paraphrases with modified attributes or properties, 
focusing on the setting without parallel data~\citep{hu-1,shen-1}. This setting poses challenges for evaluation. We show that the metric of post-transfer classification accuracy is insufficient on its own, and propose additional metrics based on semantic preservation and fluency as well as a way to combine them into a single overall score. 
We contribute new loss functions and training strategies to address the different metrics. 
Semantic preservation is addressed by adding a cyclic consistency loss and a loss based on paraphrase pairs, 
while fluency is improved by integrating losses based on style-specific language models. 
We experiment with a Yelp sentiment dataset and a new literature dataset that we propose, using multiple models that extend prior work \citep{shen-1}. We demonstrate that our metrics correlate well with human judgments, at both the  sentence-level and system-level. Automatic and manual evaluation also show large improvements over the baseline method of \citet{shen-1}. We hope that our proposed metrics can speed up system development for new textual transfer tasks while also encouraging the 
community to address our three complementary aspects of transfer quality. 
\end{abstract}

\section{Introduction}\label{sec:intro}

We consider \textbf{textual transfer}, which we define as the capability of generating textual paraphrases with modified attributes or stylistic properties, such as 
politeness~\citep{politeness}, sentiment~\citep{hu-1,shen-1}, and   formality~\citep{formality}. 
An effective transfer system could benefit a range of user-facing text generation applications such as dialogue \citep{ritter2011data} 
and writing assistance \citep{heidorn2000intelligent}.
It can also improve NLP systems via data augmentation and domain adaptation. 

However, one factor that makes textual transfer difficult is the lack of parallel corpora. 
Advances have been made in developing transfer methods that do not require parallel corpora (see Section~\ref{sec:related}), but issues remain with automatic evaluation metrics. \citet{simple-transfer} used crowdsourcing to obtain manually-written references and used 
BLEU~\citep{papineni2002bleu} to evaluate sentiment transfer. However, this approach is costly and difficult to scale for arbitrary textual transfer tasks.

Researchers have thus turned to \emph{unsupervised} evaluation metrics that do not require references. 
The most widely-used unsupervised evaluation  
uses a pretrained style classifier and computes the fraction of times the classifier was convinced of transferred style~\citep{shen-1}. However, relying solely on this metric 
leads to models that completely distort the semantic content of the input sentence. 
Table~\ref{tab1} illustrates this tendency. 

We address this deficiency by 
identifying two competing goals: preserving semantic content and producing fluent output. We contribute two corresponding metrics. 
Since the metrics are unsupervised, they 
can be used directly for tuning and model selection, 
even on test data.
The three metric categories are complementary and help us avoid degenerate behavior in 
model selection. 
For particular applications, practitioners can choose the appropriate combination of our metrics to achieve the desired balance among transfer, semantic preservation, and fluency. It is often useful to summarize the three metrics into one number, which we discuss in Section~\ref{sec:gm}.

We also add learning criteria to the framework of \citet{shen-1} to accord with our new metrics. 
We encourage semantic preservation by adding a ``cyclic consistency'' loss (to ensure that transfer is reversible) and a loss based on 
paraphrase pairs (to show the model examples of content-preserving transformations). 
To encourage fluent outputs, we add losses based on pretrained corpus-specific language models. 
We also experiment with multiple, complementary discriminators and find that they improve the trade-off between post-transfer accuracy and semantic preservation.

To demonstrate the effectiveness of our metrics, we experiment with textual transfer models discussed above, using both their Yelp polarity dataset and a new literature dataset that we propose. Across model variants, 
our metrics 
correlate well with human judgments, at both the sentence-level and system-level.

\section{Related Work}\label{sec:related}

\paragraph{Textual Transfer Evaluation}

Recent work has included human evaluation of the three categories (post-transfer style accuracy, semantic preservation, fluency), but does not propose automatic evaluation metrics for all three~\citep{simple-transfer,back-translation,chen2018adversarial,zhang2018learning}.
There have been recent proposals for supervised evaluation metrics~\citep{simple-transfer}, but these require annotation and are therefore unavailable for new textual transfer tasks. 
There is a great deal of recent 
work in textual transfer \citep{yang2018unsupervised,santos2018fighting, zhang2018learning, logeswaran2018content, nikolov2018large}, but all either lack certain categories of unsupervised metric or lack human validation of them, which we contribute. 
Moreover, the textual transfer community lacks discussion of early stopping criteria and methods of holistic model comparison. We propose a one-number summary for transfer quality, which can be used to select and compare models. 

In contemporaneous work, \citet{mir2019evaluating} similarly proposed three types of metrics for style transfer tasks. There are two main differences compared to our work: (1) They use a style-keyword masking procedure before evaluating semantic similarity, which works on the Yelp dataset (the only dataset \citet{mir2019evaluating} test on) but does not work on our Literature dataset or similarly complicated tasks, because the masking procedure goes against preserving content-specific non-style-related words. (2) They do not provide a way of aggregating three metrics for the purpose of model selection and overall comparison. We address these two problems, and we also propose metrics that are simple in addition to being effective, which is beneficial for ease of use and widespread adoption.

\paragraph{Textual Transfer Models}

In terms of generating the transferred sentences, to address the lack of parallel data, \citet{hu-1} used variational autoencoders 
to generate content representations devoid of style, which can be converted to sentences with a specific style. \citet{goldberg-1} used conditional language models to generate sentences where the desired content and style are conditioning contexts. \citet{simple-transfer} used a feature-based approach that deletes characteristic words from the original sentence, retrieves similar sentences in the target corpus, 
and generates based on the original sentence and the characteristic words from the retrieved sentences. 
\citet{cycle-reinforce} integrated reinforcement learning into the textual transfer problem. 
Another way to address the lack of parallel data is to use learning frameworks based on adversarial objectives~\citep{gan}; several have done so for 
textual transfer~\citep{yu-1,li-1,yang-1,shen-1, fu-1}. Recent work uses target-domain language models as discriminators to provide more stable feedback in learning~\citep{yang2018unsupervised}. 

To preserve semantics more explicitly, \citet{fu-1} use a multi-decoder model to learn content representations that do not reflect styles. \citet{authorship} use a cycle constraint that penalizes $L_1$ distance between input and round-trip transfer reconstruction. Our cycle consistency loss is inspired by \citet{authorship}, together with the idea of back translation in unsupervised neural machine translation \citep{back-translation-nmt, back-translation-nmt-2}, and the idea of cycle constraints in image generation by \citet{zhu-1}. 

\section{Evaluation}\label{sec:evaluations}

\subsection{Issues with Most Existing Methods}\label{sec:problem}

\begin{table}[t]
\setlength{\tabcolsep}{4pt}
\centering
\small
\begin{tabular}{rccp{4.9cm}}
\toprule
\#ep & $\acc$ & $\gsim$ & Sentence\\
\midrule
\multicolumn{3}{c}{original input} & the host that walked us to the table and left without a word . \\
\specialrule{.2pt}{1pt}{1pt}
0.5 & 0.87 & 0.65 & the food is the best and the food is the . \\
\specialrule{.2pt}{1pt}{1pt}
3.3 & 0.72 & 0.75 & the owner that went to to the table and made a smile . \\
\specialrule{.2pt}{1pt}{1pt}
7.5 & 0.58 & 0.81 & the host that walked through to the table and are quite perfect ! \\
\bottomrule
\end{tabular}
\caption{Examples showing why $\acc$ is insufficient. The original sentence has negative sentiment, and the goal is to transfer to positive. \#ep is number of epochs trained when generating the sentence and $\gsim$ (described below) is the semantic similarity to the original sentence. 
High $\acc$ is associated with low $\gsim$. 
}\label{tab1}
\end{table}

\noindent
Prior work in automatic evaluation of textual transfer has focused on post-transfer classification accuracy (``$\acc$''), computed by using a pretrained classifier to measure classification accuracy of transferred texts~\citep{hu-1,shen-1}. However, there is a problem with relying solely on this metric. Table~\ref{tab1} shows examples of transferred sentences at several points in training the model of \citet{shen-1}.
$\acc$ is highest very early in training and decreases over time as the outputs become a stronger semantic match to the input, 
a trend we show in more detail in Section~\ref{sec:result}.
Thus transfer quality is inversely proportional to semantic similarity to the input sentence, meaning that these metrics are complementary and difficult to optimize simultaneously. 

We also identify a third category of metric, namely fluency of the transferred sentence, and similarly find it to be complementary to the first two. 
These three metrics can be used to evaluate textual transfer systems and to do hyperparameter tuning and early stopping. In our experiments, we found that training typically converges to a point that gives poor $\acc$.
Intermediate results are much better under a combination of all three unsupervised metrics. Stopping criteria are rarely discussed in prior work on textual transfer.

\subsection{Unsupervised Evaluation Metrics}\label{sec:eval}

We now describe our proposals.
We validate the metrics with human judgments in Section~\ref{sec:metrics-validation}. 
\vspace{-0.15cm}
\paragraph{Post-transfer classification accuracy (``$\acc$''):} This metric was mentioned above. We use a CNN \citep{kim-1} trained to classify a sentence as being 
from $\mathbf{X}_0$ or $\mathbf{X}_1$ (two corpora corresponding to different styles or attributes). Then $\acc$ is the percentage of transferred sentences that are classified as belonging to the transferred class. 
\vspace{-0.15cm}
\paragraph{Semantic Similarity (``$\gsim$''):} 
We compute semantic similarity between the input and transferred sentences. 
We~embed sentences by averaging their word embeddings 
weighted by idf scores, where $\mathrm{idf}(q)=\log ({|C|}\cdot {|\{s \in C: q \in s\}|}^{-1})$ ($q$ is a word, $s$ is a sentence, $C=\mathbf{X}_0 \cup \mathbf{X}_1$). 
We use $300$-dimensional GloVe word embeddings~\citep{glove}. 
Then, $\gsim$ is the average of the cosine similarities over all original/transferred sentence pairs. 
Though this metric is quite simple, 
we show empirically that it is effective in capturing semantic similarity. 
Simplicity in evaluation metrics is beneficial for computational efficiency and widespread adoption. The quality of transfer evaluations will be significantly boosted with even such a simple metric. We also experimented with METEOR \citep{denkowski-1}. However, given that we found it to be strongly correlated with $\gsim$ (shown in supplemental materials), we adopt $\gsim$ due to its computational efficiency and simplicity.

Different textual transfer tasks may require different degrees of semantic preservation. 
Our summary metric, described in Section~\ref{sec:gm}, can be tailored by practitioners for various datasets and tasks which may require more or less weight on semantic preservation. 

\vspace{-0.15cm}
\paragraph{Fluency (``$\pp$''):} Transferred sentences can exhibit high $\acc$ and $\gsim$ while still being ungrammatical. So we add a third unsupervised metric to target fluency. 
We compute perplexity (``$\pp$'') of the transferred corpus, using a language model pretrained on the concatenation of $\mathbf{X}_0$ and $ \mathbf{X}_1$. 
We note that perplexity is distinct from fluency. However, certain measures based on perplexity have been shown to correlate with sentence-level human fluency judgments \citep{gamon2005sentence,DBLP:conf/conll/KannRF18}. 
Furthermore, as discussed in Section~\ref{sec:gm}, we punish abnormally small perplexities, as transferred texts with such perplexities typically consist entirely of words and phrases that do not result in meaningful sentences. 
Our summary metric, described in Section~\ref{sec:gm}, can be tailored by practitioners for various datasets and tasks which may require more or less weight on semantic preservation.

\subsection{Summarizing Metrics into One Score} \label{sec:gm}
\label{sec:new-sec-improve}

It is often useful to summarize multiple metrics into one number, for ease of tuning and model selection. To do so, we propose an adjusted geometric mean ($\gm$) of a generated sentence $q$:

\vspace*{-6mm}
\begin{align} 
& \gm_\mathbf{t}(q) = \big( [\mathrm{100\cdot \acc} - t_1]_+ \cdot [\mathrm{100\cdot \gsim} - t_2]_+ \nonumber\\
&\quad \cdot \min \{ [t_3-\pp]_+, [\pp-t_4]_+ \} \big)^{\frac{1}{3}} \label{eqn:gm}
\end{align}
where $\mathbf{t}=(t_i)_{i \in [4]}$, and $[\cdot]_+ = \max(\cdot,0)$. 
Note that as discussed above, we punish abnormally small perplexities by setting $t_4$.

When choosing models, different practitioners may prefer different trade-offs of $\acc$, $\gsim$, and $\pp$. 
As one example, we provide a set~of parameters based on \textit{our} experiments: 
$\mathbf{t}=(63, 71, 97, -37)$.
We sampled 300 pairs of transferred sentences from a range of models from our two different tasks 
(Yelp and literature) 
and asked annotators which of the two sentences is better.
We denote a pair of sentences by $(y^+, y^-)$ where $y^+$ is preferred. We train the parameters $\mathbf{t}$ 
using the following loss:
\begin{equation}
L_\gm(\mathbf{t}) = \max (0, -\gm_\mathbf{t}(y^+)+\gm_\mathbf{t}(y^-)+1)\nonumber
\end{equation}
\noindent 
In future work, a richer function 
$f(\acc, \gsim, \pp)$ 
could be learned from additional annotated data, and more diverse textual transfer tasks can be integrated into the parameter training.

\section{Textual Transfer Models}\label{sec:model}
\label{sec:losses}

The textual transfer systems introduced below are designed to target the metrics. These system variants are also used for metric evaluation. 
Note that each variant of the textual transfer system uses different components described below. 

Our model is based on \citet{shen-1}. 
We define 
$\mathbf{y}\in \mathbb{R}^{200}$ and $\mathbf{z}\in \mathbb{R}^{500}$ to be latent style
and content variables, respectively. 
$\mathbf{X}_0$ and $\mathbf{X}_1$ are two corpora containing sentences $\mathbf{x}_0^{(i)}$ and $\mathbf{x}_1^{(i)}$ respectively, where the word embeddings are in $\mathbb{R}^{100}$. 
We transfer using an encoder-decoder framework. The encoder $E:\mathcal{X}\times \mathcal{Y} \rightarrow \mathcal{Z}$ (where $\mathcal{X}, \mathcal{Y}, \mathcal{Z}$ are sentence domain, style space, and content space, respectively) is defined using an RNN with gated recurrent unit (GRU; \citealp{chung2014empirical}) cells. 
The decoder/generator $G:\mathcal{Y}\times \mathcal{Z} \rightarrow \mathcal{X}$ is defined also using a GRU RNN. 
We use $\widetilde{\mathbf{x}}$ to denote the style-transferred version of $\mathbf{x}$. We want $\textstyle{{\widetilde{\mathbf{x}}_t^{(i)}} = G(\mathbf{y}_{1-t},E(\mathbf{{x}}_t^{(i)},\mathbf{y}_t))}$ for $t\in\{0, 1\}$.

\subsection{Reconstruction and Adversarial Losses}\label{sec:recadv}

\citet{shen-1} used two families of losses for training: 
reconstruction and adversarial losses. The reconstruction loss solely helps the encoder and decoder work well at encoding and generating natural language, without any attempt at transfer: 
\vspace*{-3mm}
\begin{multline}\label{eqn:cons}
\lrec(\mathbf{\theta}_E, \mathbf{\theta}_G)
\\
=\textstyle{\sum_{t=0}^1 \mathbb{E}_{\mathbf{x}_t} \big[-\log p_G (\mathbf{x}_t \mid \mathbf{y}_t, E({\mathbf{x}_t},\mathbf{y}_t))\big]}
\end{multline}
\noindent 
The loss seeks to ensure that when a sentence $\mathbf{x}_t$ is encoded to its content vector and then decoded to generate a sentence, the generated sentence should match $\mathbf{x}_t$. 
For their adversarial loss, \citet{shen-1} used a pair of discriminators: $D_0$ tries to distinguish between $\mathbf{x}_0$ and $\widetilde{\mathbf{x}}_1$, and $D_1$ between $\mathbf{x}_1$ and $\widetilde{\mathbf{x}}_0$. 
In particular, decoder $G$'s hidden states are aligned instead of output words.
\vspace*{-3mm}
\begin{multline} \label{eqn:adv}
\ladv{t}(\mathbf{\theta}_E, \mathbf{\theta}_G, \mathbf{\theta}_{D_t})= 
\textstyle{-\frac{1}{k} \sum_{i=1}^k \log D_t (\mathbf{h}_t^{(i)})}\\
\textstyle{-\frac{1}{k} \sum_{i=1}^k \log (1 - D_t (\widetilde{\mathbf{h}}_{1-t}^{(i)}))}
\end{multline}
where $k$ is the size of a mini-batch. 
$D_t$ outputs the probability that its input is from style $t$ 
where the classifiers are based on the convolutional neural network from \citet{kim-1}. The CNNs use filter $n$-gram sizes of 3, 4, and 5, with 128 filters each. We obtain hidden states $\mathbf{h}$ by unfolding $G$ from the initial state $(\mathbf{y}_t, \mathbf{z}_t^{(i)})$ and feeding in $\mathbf{x}_t^{(i)}$. We obtain hidden states $\widetilde{\mathbf{h}}$ by unfolding $G$ from $(\mathbf{y}_{1-t}, \mathbf{z}_t^{(i)})$ and feeding in the previous output probability distributions.

\subsection{Cyclic Consistency Loss}\label{sec:cyc}

We use a ``cyclic consistency'' loss \citep{zhu-1} to encourage already-transferred sentences to be able to be recovered by transferring back again. This loss is similar to $\lrec$ except we now transfer style twice in the loss. 
Recall that we seek to transfer $\mathbf{x}_t$ to $\widetilde{\mathbf{x}}_t$. 
After successful transfer, we expect $\widetilde{\mathbf{x}}_t$ to have style $\mathbf{y}_{1-t}$, 
and $\widetilde{\widetilde{\mathbf{x}}}_t$ (transferred back from $\widetilde{\mathbf{x}}_t$) to have style $\mathbf{y}_t$. 
We want $\widetilde{\widetilde{\mathbf{x}}}_t$ to be very close to the original untransferred $\mathbf{x}_t$. The loss is defined as 
\vspace*{0mm}
\begin{multline} \label{eqn:cyc}
\textstyle{\!\!\!\!\!\!\lcyc(\mathbf{\theta}_E, \mathbf{\theta}_G)}
\textstyle{=\!\!\sum_{t=0}^1 \mathbb{E}_{\mathbf{x}_t} \!\big[\!\!-\!\!\log p_G ({{\mathbf{x}_t}} \!\mid\! \mathbf{y}_t, \widetilde{\mathbf{z}}_t) \big]}
\end{multline}
where $\widetilde{\mathbf{z}}_t = E(G(\mathbf{y}_{1-t},E(\mathbf{{x}}_t,\mathbf{y}_t)), \mathbf{y}_{1-t})$ or, more concisely, 
$\widetilde{\mathbf{z}}_t = E(\widetilde{\mathbf{x}}_t,\mathbf{y}_{1-t})$.

To use this loss, the first step is to transfer sentences $\mathbf{x}_{t}$ from style $t$ to ${1-t}$ to get ${\widetilde{\mathbf{x}}_{t}}$. The second step is to transfer ${\widetilde{\mathbf{x}}_t}$ of style ${1-t}$ back to $t$ so that we can compute the loss of the words in $\mathbf{x}_t$ using probability distributions computed by the decoder. 
Backpropagation on the embedding, encoder, and decoder parameters will only be based on the second step, because the first step involves argmax operations which prevent backpropagation. Still, we find that the cyclic loss greatly improves semantic preservation during transfer.

\subsection{Paraphrase Loss}\label{sec:para}
While $\lrec$ provides the model with one way to preserve style (i.e., simply reproduce the input), the model does not see any examples of style-preserving paraphrases. 
To address this, we add a paraphrase loss very similar to losses used in neural machine translation. 
We define the loss on a sentential paraphrase pair $\langle\mathbf{u}, \mathbf{v}\rangle$ and assume that $\mathbf{u}$ and $\mathbf{v}$ have the same style and content. 
The loss is the sum of token-level log losses for generating each word in $\mathbf{v}$ conditioned on the encoding of $\mathbf{u}$: 
\begin{align} 
&\textstyle{\lpara(\theta_E, \theta_G)}\nonumber\\
&\textstyle{=\sum_{t=0}^1\textstyle{\mathbb{E}_{\langle\mathbf{u}, \mathbf{v}\rangle} \big[\!-\!\log p_G (\mathbf{v} \mid \mathbf{y}_t, E(\mathbf{u},\mathbf{y}_t))\big]}} \label{eqn:tl}
\end{align}
For paraphrase pairs, we use the
ParaNMT-50M dataset~\citep{wieting-1}.\footnote{We first filter out sentence pairs where one sentence is the substring of another, and then randomly select 90K pairs.}

\subsection{Language Modeling Loss}\label{sec:lang}

We attempt to improve fluency (our third metric) 
and assist transfer with a loss based on matching a pretrained language model for the target style. The loss is the cross entropy (CE) between the probability distribution from this language model and the distribution from the decoder: 
\begin{align} 
\!\!\!\textstyle{\llang(\mathbf{\theta}_E, \mathbf{\theta}_G) \!=\!\sum_{t=0}^1} \textstyle{ \mathbb{E}_{\mathbf{x}_t} \big[\sum_i \!\mathrm{CE} (\mathbf{l}_{t,i}, \mathbf{g}_{t,i})\big]} \label{eqn:lang}
\end{align}
where 
$\mathbf{l}_{t,i}$ and $\mathbf{g}_{t,i}$ are distributions over the vocabulary defined as follows:
\vspace*{-2mm}
\begin{align}
\mathbf{l}_{t,i} &=p_{\mathit{LM}_{1-t}} (\,\cdot \mid \widetilde{\mathbf{x}}_{t_{1:(i-1)}})\nonumber\\
\mathbf{g}_{t,i} &= p_{G} (\,\cdot \mid \widetilde{\mathbf{x}}_{t_{1:(i-1)}}, \mathbf{y}_{1-t},  E(\mathbf{x}_t, \mathbf{y}_{t}))\nonumber
\end{align}
where $\cdot$ stands for all 
words in the vocabulary built from the corpora. 
When transferring from style $t$ to $1-t$, 
$\mathbf{l}_{t,i}$ is the distribution 
under the language model $p_{\mathit{LM}_{1-t}}$ pretrained on sentences from style $1-t$ and 
$\mathbf{g}_{t,i}$ is the distribution  
under the decoder $G$. 
The two distributions $\mathbf{l}_{t,i}$ and $\mathbf{g}_{t,i}$ are over words at position $i$ given the $i-1$ words already predicted by the decoder. 
The two style-specific language models are pretrained on the corpora corresponding to the two styles. 
They are GRU RNNs with a dropout probability of 0.5, and they are kept fixed during the training of the transfer network.

\subsection{Multiple Discriminators}

Note that each of the textual transfer system variants uses different losses or components described in this section. To create more variants,
we add a second pair of discriminators, $D_0'$ and $D_1'$, to the adversarial loss to address the possible mode collapse problem~\citep{dual-d}. In particular, we use CNNs with $n$-gram filter sizes of 3, 4, and 5 for $D_0$ and $D_1$, and we use CNNs with $n$-gram sizes of 1, 2, and 3 for $D_0'$ and $D_1'$. Also, for $D_0'$ and $D_1'$, we use the Wasserstein GAN  (WGAN) framework~\citep{wgan}. 
The adversarial loss takes the following form:
\vspace*{-3mm}
\begin{multline} \label{eqn:advp}
\ladvp{t}(\mathbf{\theta}_E, \mathbf{\theta}_G, \mathbf{\theta}_{D_t'})= \textstyle{\frac{1}{k} \sum_{i=1}^k \big[ D_t'(\widetilde{\mathbf{h}}_t^{(i)})} \\
\textstyle{-D_t'(\mathbf{h}_t^{(i)}) + \xi (\norm{\nabla_{\hcirc_t^{(i)}} D_t'(\hcirc_t^{(i)})}_2 - 1)^2 }\big]
\end{multline}
where $\hcirc_t^{(i)} = \epsilon_i \mathbf{h}_t^{(i)} + (1-\epsilon_i) \widetilde{\mathbf{h}}_t^{(i)}$ where $\epsilon_i \sim \mathrm{Uniform}([0,1])$ is sampled for each training instance. The adversarial loss is based on \citet{wgan},\footnote{We use a default value of $\xi=10$.} with the exception that we use the hidden states of the decoder instead of word distributions as inputs to $D_t'$, similar to Eq.~(\ref{eqn:adv}).

We choose WGAN 
in the hope that its differentiability properties
can help avoid vanishing gradient and mode collapse problems. 
We expect the generator to receive helpful gradients even if the discriminators perform well. 
This approach leads to much better outputs, as shown below. 

\subsection{Summary}\label{sec:training-summary}

We iteratively update (1) $\theta_{D_0}$, $\theta_{D_1}$, $\theta_{D'_0}$, and $\theta_{D'_1}$ by gradient descent on $\ladv{0}$, $\ladv{1}$, $\ladvp{0}$, and $\ladvp{1}$, respectively, and (2) $\theta_E$, $\theta_G$ by gradient descent on $L_{\mathit{total}} = \lambda_1 \lrec + \lambda_2 \lpara + \lambda_3 \lcyc + \lambda_4 \llang - \lambda_5 (\ladv{0} + \ladv{1})  - \lambda_6 (\ladvp{0} + \ladvp{1})$. Depending on which model is being trained (see Table \ref{table:result-1}), the $\lambda_i$'s for the unused losses will be zero. More details are shown in Section~\ref{sec:experiment}. 
The appendix shows the full algorithm.

\section{Experimental Setup}\label{sec:experiment}

\subsection{Datasets}\label{sec:data}

\paragraph{Yelp sentiment.} We use the same Yelp dataset as \citet{shen-1}, which uses corpora of positive and negative Yelp reviews. The goal of the transfer task is to generate rewritten sentences with similar content but inverted sentiment. We use the same train/development/test split as \citet{shen-1}. 
The dataset has 268K, 38K, 76K positive training, development, and test sentences, respectively, and 179K/25K/51K negative sentences. 
Like \citet{shen-1}, we only use sentences with 15 or fewer words.  

\paragraph{Literature.} We consider two corpora of literature. The first corpus contains works of Charles Dickens collected from Project Gutenberg. 
The second corpus is comprised of modern literature from the Toronto Books Corpus~\citep{toronto-book}. Sentences longer than 25 words are removed. 
Unlike the Yelp dataset, the two corpora have very different vocabularies. This dataset poses challenges for the textual transfer task, and it provides diverse data for assessing quality of our evaluation system. 
Given the different and sizable vocabulary, we preprocess by using the named entity recognizer in Stanford CoreNLP \citep{manning2014stanford} to replace names and locations with \textsc{-person-} and \textsc{-location-} tags, respectively. We also use byte-pair encoding (BPE), commonly used in generation tasks~\citep{P16-1162}. We only use sentences with lengths between 6 and 25. The resulting dataset has 156K, 5K, 5K Dickens training, development, and testing sentences, respectively, and 165K/5K/5K modern literature sentences.

\subsection{Hyperparameter Settings}

Section~\ref{sec:training-summary} requires setting the $\lambda$ weights for each component. Depending on which model is being trained (see Table \ref{table:result-1}), the $\lambda_i$'s for the unused losses will be zero. Otherwise, we set $\lambda_1 = 1$, $\lambda_2 = 0.2$, $\lambda_3 = 5$, $\lambda_4 = 10^{-3}$, $\lambda_5 = 1$, $\lambda_6 = 2^{-ep}$ where $ep$ is the number of epochs. 
For optimization we use Adam~\citep{adam} with a learning rate of $10^{-4}$. We implement our models using TensorFlow \citep{tf-2015}.\footnote{Our implementation is based on code from \citet{shen-1}.} Code will be available via the first author's webpage \texttt{yzpang.me}.

\subsection{Pretrained Evaluation Models} 
For the pretrained classifiers, the accuracies on the Yelp and Literature development sets are 0.974 and 0.933, respectively.
For language models, the perplexities on the Yelp and Literature development sets are 27.4 and 40.8, respectively.

\section{Results and Analysis} 
\label{sec:new-sec-human}
\label{sec:result}

\begin{table}[t]
\centering
\resizebox{\linewidth}{!}{%
\small
\centering
\begin{tabular}{lrcccccc}%cccc}
\toprule
    & & $\acc$ & $\gsim$ & $\pp$ & $\gm$ \\
\midrule
   M0: \citet{shen-1} & & 0.818 & 0.719 & 37.3 & 10.0 \\
   M1: M0\textit{+para} & & 0.819 & 0.734 & 26.3 & 14.2 \\
   M2: M0\textit{+cyc} & & 0.813 & 0.770 & 36.4 & 18.8 \\
   M3: M0\textit{+cyc+lang} & & 0.807 & 0.796 & 28.4 & 21.5 \\
   M4: M0\textit{+cyc+para} & & 0.798 & 0.783 & 39.7 & 19.2 \\
   {M5: M0\textit{+cyc+para+lang}} & & 0.804 & 0.785 & 27.1 & 20.3 \\
   {M6: M0\textit{+cyc+2d}} & & 0.805 & \textbf{0.817} & 43.3 & 21.6 \\
   {M7: M6+\textit{para+lang}} & & 0.818 & {0.805} & \textbf{29.0} & \textbf{22.8} \\
\bottomrule
\end{tabular}%
}
\caption{Yelp results with various systems and automatic metrics
at a nearly-fixed $\acc$, with best scores in boldface. We use M0 to denote \citet{shen-1}. 
\label{table:result-1}
}
\end{table}

\begin{table}[t]
\centering
\resizebox{\linewidth}{!}{%
\small
\centering
\begin{tabular}{lrcccccc}
\toprule
    & & $\acc$ & $\gsim$ & $\pp$ & $\gm$ \\
\midrule
   M0: \citet{shen-1} & & 0.694 & 0.728 & \textbf{22.3} & 8.81 \\
   M1: M0\textit{+para} & & 0.702 & 0.747 & 23.6 & 11.7 \\
   M2: M0\textit{+cyc} & & 0.692 & 0.781 & 49.9 & \textbf{12.8} \\
   M3: M0\textit{+cyc+lang} & & 0.698 & 0.754 & 39.2 & 12.0 \\
   M4: M0\textit{+cyc+para} & & 0.702 & 0.757 & 33.9 & \textbf{12.8} \\
   {M5: M0\textit{+cyc+para+lang}} & & 0.688 & 0.753 & 28.6 & 11.8 \\
   {M6: M0\textit{+cyc+2d}} & & 0.704 & \textbf{0.794} & 63.2 & \textbf{12.8} \\
   {M7: M6+\textit{para+lang}} & & 0.706 & {0.768} & 49.0 & \textbf{12.8} \\
\bottomrule
\end{tabular}%
}
\caption{Literature results with various systems and automatic metrics 
at a nearly-fixed $\acc$, with best scores in boldface. We use M0 to denote \citet{shen-1}.
\label{table:result-2}
}
\end{table}

\begin{figure}[h!]
    \centering
    \begin{subfigure}[b]{0.46\textwidth}
        \centering
        \includegraphics[width=\columnwidth]{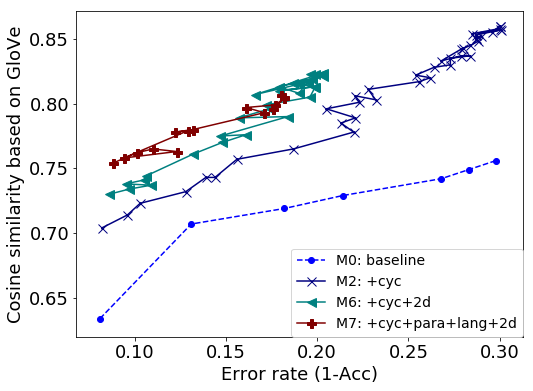}
    \end{subfigure}
    \\
    \begin{subfigure}[b]{0.44\textwidth}  
        \centering 
        \includegraphics[width=\columnwidth]{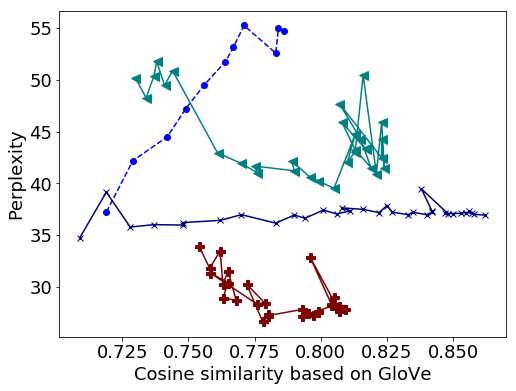}
    \end{subfigure}
    \caption{Learning trajectories 
    with models from Table \ref{table:result-1}. Metrics
    are computed on the dev sets. Figures for Literature (with similar trends) are in supplementary.\label{fig0}}
\end{figure}

\begin{table*}[t]
\setlength{\tabcolsep}{5pt}
\centering
\small
\begin{tabular}{cccccccccccc|ccccc|c}
\toprule
 & & \multicolumn{2}{c}{Models} & & \multicolumn{3}{c}{Transfer quality} & & \multicolumn{4}{c}{Semantic preservation} & & \multicolumn{4}{c}{Fluency}\\
 \cline{3-4} \cline{6-8} \cline{10-14} \cline{15-18} \noalign{\smallskip}
Dataset & & A & B & & A$>$B & B$>$A & Tie & & A$>$B & B$>$A & Tie & $\Delta_{\gsim}$ & & A$>$B & B$>$A & Tie & $\Delta_{\pp}$\\
\cline{1-1} \cline{3-4} \cline{6-8} \cline{10-13} \cline{15-18} \noalign{\smallskip}
 & & M0 & M2 & & 9.0 & 6.0 & 85.1 & & 1.5 & \textbf{25.4} & 73.1 & -0.05 & & 10.4 & \textbf{23.9} & 65.7 & 0.9\\
\multirow{2}{*}{Yelp} & & M0 & M7 & & 9.6 & 14.7 & 75.8 & & 2.5 & \textbf{54.5} & 42.9 & -0.09 & & 4.6 & \textbf{39.4} & 56.1 & 8.3\\
 & & M6 & M7 & & 13.7 & 11.6 & 74.7 & & 16.0 & 16.7 & 67.4 & 0.01 & & 10.3 & {20.0} & 69.7 & 14.3\\
 & & M2 & M7 & & 5.8 & 9.3 & 84.9 & & 8.1 & \textbf{25.6} & 66.3 & -0.04 & & 14.0 & \textbf{26.7} & 59.3 & 7.4\\
\cline{1-1} \cline{3-4} \cline{6-8} \cline{10-13} \cline{15-18} \noalign{\smallskip}
\multirow{2}{*}{Literature} & & M2 & M6 & & 4.2 & 6.7 & 89.2 & & {16.7} & 20.8 & 62.5 & 0.01 & & \textbf{40.8} & 13.3 & 45.8 & -13.3\\
 & & M6 & M7 & & 15.8 & 13.3 & 70.8 & & \textbf{25.0} & 9.2 & 65.8 & 0.03 & & 14.2 & {20.8} & 65.0 & 14.2\\

\bottomrule
\end{tabular}
\caption{Manual evaluation results (\%) using models from 
Table~\ref{table:result-1} (i.e., with roughly fixed $\acc$). 
$>$ means ``better than''. $\Delta_{\gsim}=\gsim(A)-\gsim(B)$, and $\Delta_{\pp}=\pp(A)-\pp(B)$ (note that lower $\pp$ generally means better fluency). Each row uses at least 120 sentence pairs. 
A cell is bold if it represents a model win of at least 10\%. 
\label{table:human}} 
\end{table*}

\subsection{Analyzing Metric Relationships}
\label{sec:relationship}
Table~\ref{table:result-1} shows results for the Yelp dataset and Figure~\ref{fig0} plots learning trajectories of those models. Table~\ref{table:result-2} shows results for the Literature dataset. 
Models for the Literature dataset show similar trends. 
The figures show trajectories of statistics on corpora transferred/generated from the dev set during learning. Each two consecutive markers deviate by half an epoch of training. 
Lower-left markers generally precede upper-right ones.
In Figure~\ref{fig0}(a), the plots of $\gsim$ by error rate 
($1-\acc$) exhibit positive slopes, meaning that error rate 
is positively correlated with $\gsim$. Curves to the upper-left corner represent better trade-off between error rate and $\gsim$. 
In the plots of $\pp$ by $\gsim$ in Figure~\ref{fig0}(b), the M0 curve exhibits large positive slope but the curves for other models do not, which indicates that M0 sacrifices $\pp$ for $\gsim$. 
Other models maintain consistent $\pp$ as $\gsim$ increases during training.

\subsection{System-Level Validation}
\label{sec:human}

Annotators were shown the 
untransferred sentence, as well as sentences produced by two models (which we refer to as A and B). They were asked to judge which better reflects the target style (A, B, or tie), which has better semantic preservation of the original (A, B, or tie), and which is more fluent (A, B, or tie). 
Results 
are shown in Table~\ref{table:human}. 

Overall, the results show the same trends as our automatic %evaluation 
metrics. 
For example, on Yelp, 
large differences in human judgments of semantic preservation (M2$>$M0, M7$>$M0, M7$>$M2) also show the largest differences in $\gsim$, while M6 and M7 have very similar human judgments and very similar $\gsim$ scores.

\subsection{Sentence-Level Validation of Metrics}
\label{sec:metrics-validation}

\begin{table}[t]
\setlength{\tabcolsep}{4pt}
\centering
\small
\begin{tabular}{cp{4.5cm}cc}
\toprule
Metric & Method of validation & Yelp & Lit. \\
\midrule
$\acc$ & \% of machine and human judgments that match & 94 & 84\\
\specialrule{.2pt}{1pt}{1pt}
$\gsim$ & Spearman's $\rho$ b/w $\gsim$ and human ratings of semantic preservation & 0.79 & 0.75\\
\specialrule{.2pt}{1pt}{1pt}
$\pp$ & Spearman's $\rho$ b/w negative $\pp$ and human ratings of fluency & 0.81 & 0.67\\
\bottomrule
\end{tabular}%
%}
%}
\caption{Human sentence-level validation of metrics; 100 examples for each dataset for validating $\acc$; 150 each for $\gsim$ and $\pp$; see text for validation of $\gm$. \label{tab2}}
\end{table}

We describe a human sentence-level validation of our metrics in Table~\ref{tab2}. 

To validate $\acc$, human annotators were asked to judge the style of 100 transferred sentences (sampled equally from M0, M2, M6, M7). Note that it is a binary choice question (style 0 or style 1 without ``tie'' option) so that human annotators had to make a choice. We then compute the percentage of machine and human judgments that match. 

We validate $\gsim$ and $\pp$ by computing sentence-level Spearman's $\rho$ 
between the metric and human judgments (an integer score from 1 to 4) on 150 generated sentences (sampled equally from M0, M2, M6, M7). We presented pairs of original sentences and transferred sentences to human annotators. They were asked to rate the level of semantic similarity (and similarly for fluency) where 1 means ``extremely bad'', 2 means ``bad/ok/needs improvement'', 3 means ``good'', and 4 means ``very good.'' They were also given 5 examples for each rating (i.e., a total of 20 for four levels) before annotating. From Table~\ref{tab2}, all validations show strong correlations on the Yelp dataset and reasonable correlations on Literature. 

We validate $\gm$ by obtaining human pairwise preferences (without the ``tie'' option) of overall transfer quality 
and measuring the fraction of pairs in which the $\gm$ score agrees with the human preference. 
Out of 300  
pairs (150 from each dataset), 258 (86\%) match.

The transferred sentences used in the evaluation are sampled from the development sets produced by models M0, M2, M6, and M7, at the accuracy levels used in Table~2. 
In the data preparation for the manual annotation, there is sufficient randomization regarding model and textual transfer direction.

\subsection{Comparing Losses}

\paragraph{Cyclic Consistency Loss.} 
We compare the trajectories of the baseline model (M0) and the \textit{+cyc} model (M2). Table~\ref{table:result-1} and Figure~\ref{fig0} show that under similar $\acc$, M2 has much better semantic similarity for both Yelp and Literature. 
In fact, cyclic consistency loss proves to be the strongest driver of semantic preservation across all of our model configurations. The other losses do not constrain the semantic relationship across style transfer, so we include the cyclic loss in M3 to M7.

\vspace{-0.1cm}
\paragraph{Paraphrase Loss.}
Table~\ref{table:result-1} shows that the model with paraphrase loss (M1) slightly improves $\gsim$ over M0 on both datasets under similar $\acc$. For Yelp, M1 has better $\acc$ and $\pp$ than M0 at comparable semantic similarity. 
So, when used alone, the paraphrase loss helps.  
However, when combined with other losses (e.g., compare M2 to M4), its benefits are mixed. For Yelp, M4 is slightly better in preserving semantics and producing fluent output, but for Literature, M4 is slightly worse. 
A challenge in introducing an additional paraphrase dataset is that its notions of similarity may clash with those of content preservation in the transfer task. 
For Yelp, both corpora share a great deal of semantic content, 
but Literature shows systematic semantic differences  
even after preprocessing.

%\subsubsection*{Language Modeling Loss}
\vspace{-0.1cm}
\paragraph{Language Modeling Loss.}
When comparing between M2 and M3, between M4 and M5, and between M6 and M7, we find that the addition of the language modeling loss reduces $\pp$, sometimes at a slight cost of semantic preservation.

\subsection{Results based on Supervised Evaluation}\label{sec:stanford-results}

\begin{table}[h!]
\centering
\begin{small}
{
\renewcommand{\arraystretch}{0.84}
\resizebox{4.5cm}{!}{%
\setlength{\tabcolsep}{3pt}
\begin{tabular}{lcc}
\toprule
Model & BLEU & $\acc^\ast$ \\
\midrule
\citet{fu-1}\\
\quad Multi-decoder & 7.6 & 0.792\\
\quad Style embed. & 15.4 & 0.095\\
\specialrule{.2pt}{1pt}{1pt}
\multicolumn{3}{l}{\citet{simple-transfer}}\\
\quad Template & 18.0 & 0.867\\
\quad Delete/Retrieve & 12.6 & 0.909\\
\specialrule{.2pt}{1pt}{1pt}
\multicolumn{3}{l}{\citet{yang2018unsupervised}}\\
\quad LM & 13.4 & 0.854 \\
\quad LM + classifier & \textbf{22.3} & 0.900 \\
\specialrule{.2pt}{1pt}{1pt}
Untransferred & \textbf{31.4} & 0.024\\
\bottomrule
\end{tabular}%
}
\ 
\resizebox{2.9cm}{!}{%
\setlength{\tabcolsep}{4pt}
\begin{tabular}{lcc}
\toprule
M. & BLEU & $\acc$ \\
\midrule
M0 & 4.9 & 0.818 \\
\specialrule{.2pt}{1pt}{1pt}
M6 & 22.3 & 0.804\\
M6 & \textbf{22.5} & 0.843 \\
M6 & 16.3 & 0.897\\
\specialrule{.2pt}{1pt}{1pt}
M7 & 17.0 & 0.814\\
M7 & 16.3 & 0.839\\
M7 & 12.9 & 0.901\\
\bottomrule
\end{tabular}%
}
}
\end{small}
\caption{Results on Yelp sentiment transfer, where BLEU is between 1000 transferred sentences and 
human references, and $\acc$ is restricted to the same 1000 sentences. Our best models (right table) achieve higher BLEU than prior work at similar levels of $\acc$, but untransferred sentences achieve the highest BLEU. 
$\acc^\ast$: 
the definition of $\acc$ varies by row because of different classifiers in use. 
Other results from \citet{simple-transfer} are not included as they are worse.
}\label{table:human-500}
\end{table}

If we want to compare the models using one single number, $\gm$ is our unsupervised approach.
We can also compute BLEU scores between our generated outputs and human-written gold standard outputs using the 1000 Yelp references from \citet{simple-transfer}. 
For BLEU scores reported for the methods of \citet{simple-transfer}, we use the values reported by \citet{yang2018unsupervised}. We use the same BLEU implementation as used by \citet{yang2018unsupervised}, i.e., \texttt{multi-bleu.perl}. 
We compare three models selected during training from each of our M6 and M7 settings. 
We also report post-transfer accuracies reported by prior work, as well our own computed $\acc$ scores for M0, M6, M7, and the untransferred sentences. Though the classifiers differ across models, their accuracy tends to be very high ($>0.97$), making it possible to make rough comparisons of $\acc$ across them. 

BLEU scores and post-transfer accuracies are shown in Table~\ref{table:human-500}. 
The most striking result is that \emph{untransferred} sentences have the highest BLEU score by a large margin, suggesting that prior work for this task has not yet eclipsed the trivial baseline of returning the input sentence. 
However, at similar levels of $\acc$, our models have higher BLEU scores than prior work. 
We additionally find that supervised BLEU shows a trade-off with $\acc$: 
for a single model type, higher $\acc$ 
generally corresponds to lower BLEU.

\section{Conclusion}

We proposed three kinds of 
metrics for non-parallel textual transfer, studied their relationships, and developed learning criteria to address them. 
We emphasize that all three metrics are needed to make meaningful comparisons among models. 
We expect our components to be applicable to a broad range of generation tasks. 

\section*{Acknowledgments}
We thank Karl Stratos and Zewei Chu for helpful discussions, the annotators for performing manual evaluations, and the anonymous reviewers for useful comments. We also thank Google for a faculty research award to K.~Gimpel that partially supported this research.

\clearpage

\bibliography{acl2019}
\bibliographystyle{acl_natbib}

\clearpage

\appendix

\section{Supplementary Material}

\subsection{Textual Transfer Model}

\subsubsection{Summary}

We iteratively update (1) $\theta_{D_0}$, $\theta_{D_1}$, $\theta_{D'_0}$, and $\theta_{D'_1}$ by gradient descent on $\ladv{0}$, $\ladv{1}$, $\ladvp{0}$, and $\ladvp{1}$, respectively, and (2) $\theta_E$, $\theta_G$ by gradient descent on $L_{\mathit{total}} = \lambda_1 \lrec + \lambda_2 \lpara + \lambda_3 \lcyc + \lambda_4 \llang - \lambda_5 (\ladv{0} + \ladv{1})  - \lambda_6 (\ladvp{0} + \ladvp{1})$.

\subsubsection{Full Algorithm}

Please refer to Algorithm~\ref{algo}.

\begin{algorithm}[h!]
{\fontsize{10pt}{11.4pt}\selectfont
% \KwData{this text}
% \KwResult{how to write algorithm with \LaTeX2e }
 Pretrain language models $\mathit{LM}_0$ and $\mathit{LM}_1$ to be used in language modeling loss $\llang$. \\
 
 Initialize parameters ($\theta_E, \theta_G, \theta_{D_0}, \theta_{D_1}, \theta_{D'_0}, \theta_{D'_1}$). \\
 
 \While{losses have not converged}{
  	Sample mini-batch $\{\mathbf{x}_t^{(i)}\}_{i=1}^k$ from $\mathbf{X}_t$, and obtain transferred sentences $\{\widetilde{\mathbf{x}}_t^{(i)}\}_{i=1}^k$ by running the decoder $G(\mathbf{y}_{1-t}, E({\mathbf{x}}_t, \mathbf{y}_{t}))$, for $t =0,1$. \\
	
	Get content representations $\mathbf{z}_t^{(i)}=E(\mathbf{x}_t^{(i)},\mathbf{y}_t)$,  and $\widetilde{\mathbf{z}}_t^{(i)}=E(\widetilde{\mathbf{x}}_t^{(i)},\mathbf{y}_{1-t})$ for $t = 0,1$, $\forall i$, where we use $\mathbf{x}_t^{(i)}$ as inputs for the RNNs and $\mathbf{y}_{1-t}$ as initial hidden states for the RNNs. \\
	
	Obtain probability distribution of the back-transferred sentences $\{\widetilde{\widetilde{\mathbf{x}}}_t^{(i)}\}_{i=1}^k$ through decoder $G(\mathbf{y}_{t}, E({\widetilde{\mathbf{x}}}_t, \mathbf{y}_{1-t}))$, for $t = 0,1$, $\forall i$.\\
	
	Unfold G from $(\mathbf{y}_t, \mathbf{z}_t^{(i)})$ (i.e., by using $(\mathbf{y}_t, \mathbf{z}_t^{(i)})$ as initial hidden state of the RNN), and feed in $\mathbf{x}_t^{(i)}$ to obtain $\mathbf{h}_t^{(i)}$; and unfold G from $(\mathbf{y}_{1-t}, \mathbf{z}_t^{(i)})$, and feed in previous output probability distributions to obtain $\widetilde{\mathbf{h}}_t^{(i)}$. This step is done for $t =0,1$, $\forall i$.\\
	
  Compute $\lrec$ by (1); Compute $\ladv{0}$ and $\ladv{1}$ of the first discriminator by (2), and $\ladvp{0}$ and $\ladvp{1}$ of the second discriminator by (6); Compute $\lcyc$ by (3);  Compute $\lpara$ by (4); Compute $\llang$ by (5). \\
  
  Update $\theta_{D_0}$, $\theta_{D_1}$, $\theta_{D'_0}$, and $\theta_{D'_1}$ by gradient descent on $\ladv{0}$, $\ladv{1}$, $\ladvp{0}$, and $\ladvp{1}$, respectively.\\
  
  Update $\theta_E$, $\theta_G$ by gradient descent on $L_{\mathit{total}} = \lambda_1 \lrec + \lambda_2 \lpara + \lambda_3 \lcyc + \lambda_4 \llang - \lambda_5 (\ladv{0} + \ladv{1})  - \lambda_6 (\ladvp{0} + \ladvp{1})$.
  
 }
%  \eIf{understand}{
%   go to next section\;
%   current section becomes this one\;
%   }{
%   go back to the beginning of current section\;
%  }

 \caption{Training procedure}\label{algo}
 }
\end{algorithm}

\subsection{Tables and Plots in Results}

\begin{table*}[t]
\centering
\resizebox{14.5cm}{!}{%
\renewcommand{\arraystretch}{0.84}
\centering
\begin{tabular}{lrccccccccccccc}
\toprule
 \multicolumn{1}{c}{\multirow{2}{*}{\textbf{Yelp}}} & & \multicolumn{5}{c}{$\acc \approx$ 0.800} &  &  \multicolumn{5}{c}{$\gsim \approx$ 0.800}\\
    & & $\acc (\uparrow)$ & $\gsim (\uparrow)$ & $\met (\uparrow)$ & $\pp (\downarrow)$ & $\gm (\uparrow)$ & & $\acc$ & $\gsim$ & $\met$ & $\pp$ & $\gm$ \\
   %\cline{1-1} 
   \cline{3-7} \cline{9-13} \noalign{\smallskip}
   M0: \citet{shen-1} & & 0.818 & 0.719 & 0.165 & 37.3 & 10.0 &  & 0.591 & 0.793 & 0.305 & 56.1 & 0.00 \\
   M1: M0\textit{+para} & & 0.819 & 0.734 & 0.196 & 26.3 & 14.2 &  & 0.704 & 0.798 & 0.288 & 31.0 & 16.3 \\
   M2: M0\textit{+cyc} & & 0.813 & 0.770 & 0.271 & 36.4 & 18.8 &  & 0.795 & 0.801 & 0.312 & 37.4 & 20.8\\
   M3: M0\textit{+cyc+lang} & & 0.807 & 0.796 & 0.257 & 28.4 & 21.5 &  & 0.792 & 0.802 & 0.272 & 28.7 & {21.4}\\
   M4: M0\textit{+cyc+para} & & 0.798 & 0.783 & 0.275 & 39.7 & 19.2 & & 0.794 & 0.799 & 0.320 & 39.4 & 20.3\\
   {M5: M0\textit{+cyc+para+lang}} & & 0.804 & 0.785 & 0.254 & 27.1 & 20.3 & & 0.781 & 0.794 & 0.288 & {28.0} & 20.2 \\
   {M6: M0\textit{+cyc+2d}} & & 0.805 & \textbf{0.817} & \textbf{0.322} & 43.3 & 21.6 &  & {0.834} & 0.807 & 0.321 & 47.7 & 21.4\\
   {M7: M0\textit{+cyc+para+lang+2d}} & & 0.818 & {0.805} & 0.288 & 29.0 & \textbf{22.8} & & {0.830} & 0.799 & 0.281 & \textbf{27.8} & \textbf{22.6}\\
\midrule
 \multicolumn{1}{c}{\multirow{2}{*}{\textbf{Literature}}}  & & \multicolumn{5}{c}{$\acc \approx$ 0.700} & & \multicolumn{5}{c}{$\gsim \approx$ 0.750}\\
    & & $\acc$ & $\gsim$ & $\met$ & $\pp$ & $\gm$ & & $\acc$ & $\gsim$ & $\met$ & $\pp$ & $\gm$ \\
   \cline{3-7} \cline{9-13} \noalign{\smallskip}
   M0: \citet{shen-1} & & 0.694 & 0.728 & 0.080 & 22.3 & 8.81 &  & n/a & n/a & n/a & n/a & n/a \\
   M1: M0\textit{+para} & & 0.702 & 0.747 & 0.108 & 23.6 & 11.7 &  & 0.678 & 0.749 & 0.106 & 30.8 & 10.7 \\
   M2: M0\textit{+cyc} & & 0.692 & {0.781} & 0.194 & 49.9 & \textbf{12.8} & & 0.778 & 0.754 & 0.109 & 55.0 & 14.0\\
   M3: M0\textit{+cyc+lang} & & 0.698 & 0.754 & 0.089 & 39.2 & 12.0 & & 0.698 & 0.754 & 0.089 & 39.2 & 12.0 \\
   M4: M0\textit{+cyc+para} & & 0.702 & 0.757 & 0.117 & 33.9 & \textbf{12.8} & & 0.719 & 0.756 & 0.112 & 29.7 & 14.0 \\
   {M5: M0\textit{+cyc+para+lang}} & & 0.688 & 0.753 & 0.089 & 28.6 & 11.8 & & 0.727 & 0.750 & 0.080 & \textbf{28.6} & 13.7 \\
   {M6: M0\textit{+cyc+2d}} & & 0.704 & \textbf{0.794} & \textbf{0.274} & 63.2 & \textbf{12.8} &  & 0.775 & 0.758 & 0.115 & 55.1 & \textbf{14.3} \\
   {M7: M0\textit{+cyc+para+lang+2d}} & & 0.706 & 0.768 & 0.142 & 49.0 & \textbf{12.8} & & 0.749 & 0.756 & 0.121 & 45.6 & 14.1 \\
\bottomrule
\end{tabular}%
}

\caption{Results 
at fixed levels of post-transfer classification accuracy ($\acc$) and semantic  similarity ($\gsim$). 
Under similar $\acc$, the best $\gsim$ and $\met$ are in bold. Under similar $\gsim$, the best $\pp$ is in bold. In both tables, the best $\gm$ scores are also in bold. Here, \textit{para} = paraphrase loss, \textit{cyc} = cyclic loss, \textit{lang} = language modeling loss, and \textit{2d} = two pairs of discriminators. Cells with n/a indicate that the model never reaches the corresponding $\acc$ or $\gsim$. 
\label{table:result-111}\label{table:result-222}
}
\end{table*}

Figures \ref{fig333} and \ref{fig444} show the learning trajectories for the Literature dataset, which show similar trends as those for Yelp. While the plots for the two datasets appear different from an initial glance, comparing similarities at fixed error rates and comparing perplexities at fixed similarities reveals that the results largely resemble those for the Yelp dataset. The baseline M0 struggles on the Literature dataset. The particularly low perplexity for M0 does not indicate fluent sentences, but rather the piecing together of extremely common words and phrases. 

\begin{figure}[h!]

        \centering
        \includegraphics[width=6cm]{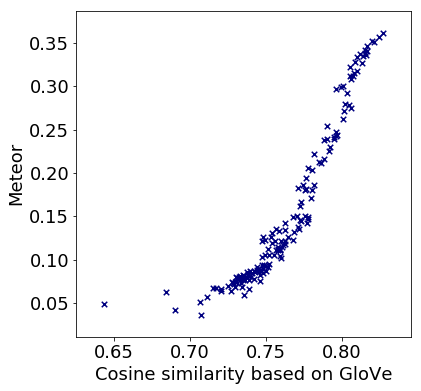}
        \caption[]%
        {{\small $\met$ by $\gsim$ using the Literature dataset}} 
        \label{fig000}
\end{figure}

In our analysis, we used $\gsim$ as the primary metric for semantic preservation. However, if we were to use $\met$ instead (where $\met$ is computed by METEOR scores between original sentence and transferred sentence, averaged over sentence pairs), the plots and our conclusions would be largely unchanged. Using the Literature dataset as an example, Figure~\ref{fig000} shows that the correlation between $\met$ and $\gsim$ is very large. Specifically, we randomly sample 200 transferred corpora generated using different models, and generated at different times during training. We obtain $\met$ and $\gsim$ of each of these 200 transferred corpora using techniques discussed in the main text. We thus have 200 data points, as shown in Figure~\ref{fig000}.

\setcounter{figure}{1}
\begin{figure*}[ht!]
    \centering
    \begin{subfigure}[b]{0.45\textwidth}   
        \centering 
        \includegraphics[width=\columnwidth]{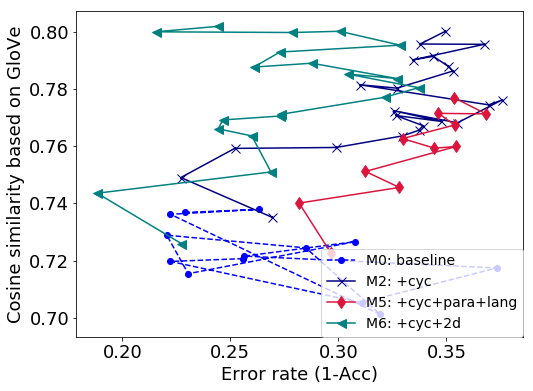}
        \caption[]%
        {{\small Cosine similarity ($\gsim$) by error rate ($1-\acc$) for Literature.}}    
        \label{fig333}
    \end{subfigure}
    \quad
    \begin{subfigure}[b]{0.45\textwidth}   
        \centering 
        \includegraphics[width=\columnwidth]{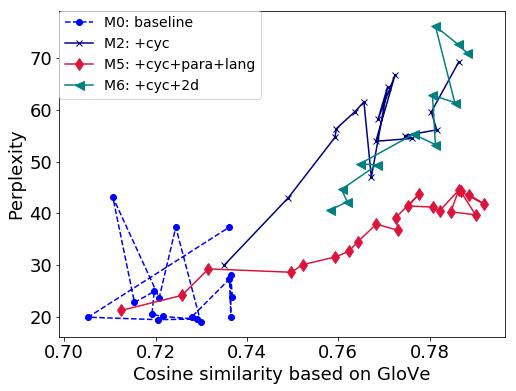}
        \caption[]%
        {{\small Perplexity ($\pp$) by cosine similarity ($\gsim$) for Literature.}}    
        \label{fig444}
    \end{subfigure}
    \caption{Learning trajectories 
    with selected models from Table 2 of main text. Metrics 
    are computed on the development sets.}
\end{figure*}

\subsection{Examples}

Table~\ref{table:ex2} provides examples of textual transfer.

\begin{table*}[h!]
\centering
\begin{small}
\begin{tabular}{ccccclc}
\toprule
Model & $\acc$ & $\gsim$ & $\pp$ & $\gm$ & Sentence & Style\\
\midrule
Original & --- & --- & --- & --- & {\makecell[l]{i got my car back and was extremely unhappy .}} & Negative\\
{M0} & 0.818 & 0.719 & 37.3 & 10.0 & {\makecell[l]{i got my favorite loves and was delicious .}} & Positive\\
{M7} & 0.818 & 0.805 & 29.0 & 22.8 & {\makecell[l]{i got my car back and was very happy .}} & Positive\\
\specialrule{.2pt}{1pt}{1pt}
Original & --- & --- & --- & --- & {\makecell[l]{the mozzarella sub is absolutely amazing .}} & Positive\\
{M0} & 0.818 & 0.719 & 37.3 & 10.0 & {\makecell[l]{the front came is not much better .}} & Negative\\
{M7} & 0.818 & 0.805 & 29.0 & 22.8 & {\makecell[l]{the cheese sandwich is absolutely awful .}} & Negative\\
\specialrule{.2pt}{1pt}{1pt}
Original & --- & --- & --- & --- & {\makecell[l]{they are completely unprofessional and have no experience .}} & Negative\\
{M0} & 0.818 & 0.719 & 37.3 & 10.0 & {\makecell[l]{they are super fresh and well !}} & Positive\\
{M7} & 0.818 & 0.805 & 29.0 & 22.8 & {\makecell[l]{they are very professional and have great service .}} & Positive\\
\specialrule{.2pt}{1pt}{1pt}
Original & --- & --- & --- & --- & {\makecell[l]{i would honestly give this place zero stars if i could .}} & Negative\\
{M0} & 0.818 & 0.719 & 37.3 & 10.0 & {\makecell[l]{i would recommend give this place from everyone again .}} & Positive\\
{M7} & 0.818 & 0.805 & 29.0 & 22.8 & {\makecell[l]{i would definitely recommend this place all stars if i could .}} & Positive\\
\specialrule{.2pt}{1pt}{1pt}
Original & --- & --- & --- & --- & {\makecell[l]{for all those reasons , we wo n't go back .}} & Negative\\
{M0} & 0.818 & 0.719 & 37.3 & 10.0 & {\makecell[l]{for all of pizza , you do you go .}} & Positive\\
{M7} & 0.818 & 0.805 & 29.0 & 22.8 & {\makecell[l]{for all those reviews , i highly recommend to go back .}} & Positive\\
\specialrule{.2pt}{1pt}{1pt}
Original & --- & --- & --- & --- & {\makecell[l]{the owner was super nice and welcoming .}} & Positive\\
{M0} & 0.818 & 0.719 & 37.3 & 10.0 & {\makecell[l]{the server was extremely bland with all .}} & Negative\\
{M7} & 0.818 & 0.805 & 29.0 & 22.8 & {\makecell[l]{the owner was very rude and unfriendly .}} & Negative\\
\specialrule{.2pt}{1pt}{1pt}
Original & --- & --- & --- & --- & {\makecell[l]{this is one of the best hidden gems in phoenix .}} & Positive\\
{M0} & 0.818 & 0.719 & 37.3 & 10.0 & {\makecell[l]{this is one of the worst \_num\_ restaurants in my life .}} & Negative\\
{M7} & 0.818 & 0.805 & 29.0 & 22.8 & {\makecell[l]{this is one of the worst restaurants in phoenix .}} & Negative\\
\specialrule{.2pt}{1pt}{1pt}
Original & --- & --- & --- & --- & {\makecell[l]{i declined on their offer , but appreciated the gesture !}} & Positive\\
{M0} & 0.818 & 0.719 & 37.3 & 10.0 & {\makecell[l]{i asked on their reviews , they are the same time !}} & Negative\\
{M7} & 0.818 & 0.805 & 29.0 & 22.8 & {\makecell[l]{i paid for the refund , and explained the frustration !}} & Negative\\

\specialrule{.2pt}{1pt}{1pt}
Original & --- & --- & --- & --- & {\makecell[l]{it was a most extraordinary circumstance .}} & Dickens\\
{M0} & 0.694 & 0.728 & 22.3 & 8.81 & {\makecell[l]{it was a little deal of the world .}} & Modern\\
{M2} & 0.692 & 0.781 & 49.9 & 12.8 & {\makecell[l]{it was a huge thing on the place .}} & Modern\\
{M6} & 0.704 & 0.794 & 63.2 & 12.8 & {\makecell[l]{it was a most important effort over the relationship .}} & Modern\\
\specialrule{.2pt}{1pt}{1pt}
Original & --- & --- & --- & --- & {\makecell[l]{i conjure you , tell me what is the matter .}} & Dickens\\
{M0} & 0.694 & 0.728 & 22.3 & 8.81 & {\makecell[l]{i 'm sorry , i 'm sure i 'm going to be , but i was a little man .}} & Modern\\
{M2} & 0.692 & 0.781 & 49.9 & 12.8 & {\makecell[l]{i 'm telling you , tell me what 's the time .}} & Modern\\
{M6} & 0.704 & 0.794 & 63.2 & 12.8 & {\makecell[l]{i am telling you , tell me what 's the matter .}} & Modern\\
\specialrule{.2pt}{1pt}{1pt}
Original & --- & --- & --- & --- & {\makecell[l]{a public table is laid in a very handsome hall for breakfast , \\ \quad and for dinner , and for supper .}} & Dickens\\
{M0} & 0.694 & 0.728 & 22.3 & 8.81 & {\makecell[l]{the other of the man was a little , and then , and -person- 's\\ \quad eyes , and then -person- .}} & Modern\\
{M2} & 0.692 & 0.781 & 49.9 & 12.8 & {\makecell[l]{a little table is standing there for all , and for me ,\\ \quad and for you .}} & Modern\\
{M6} & 0.704 & 0.794 & 63.2 & 12.8 & {\makecell[l]{a small table is placed in a very blue room for breakfast ,\\ \quad and for dinner , and for dinner .}} & Modern\\
\specialrule{.2pt}{1pt}{1pt}
Original & --- & --- & --- & --- & {\makecell[l]{does n't she know it 's dangerous for a young woman to\\ \quad go off by herself ?}} & Modern\\
{M0} & 0.694 & 0.728 & 22.3 & 8.81 & {\makecell[l]{do n't have been a little of a man of your own ?}} & Dickens\\
{M2} & 0.692 & 0.781 & 49.9 & 12.8 & {\makecell[l]{it n't she know it 's dangerous for a little woman to\\ \quad go out from us ?}} & Dickens\\
{M6} & 0.704 & 0.794 & 63.2 & 12.8 & {\makecell[l]{does n't she know it 's a dangerous act for a young lady\\ \quad to go off by herself ?}} & Dickens\\
\specialrule{.2pt}{1pt}{1pt}
Original & --- & --- & --- & --- & {\makecell[l]{it whispered to me about my new strength and abilities .}} & Modern\\
{M0} & 0.694 & 0.728 & 22.3 & 8.81 & {\makecell[l]{it is not a little man .}} & Dickens\\
{M2} & 0.692 & 0.781 & 49.9 & 12.8 & {\makecell[l]{it appears to me about my new strength and desire .}} & Dickens\\
{M6} & 0.704 & 0.794 & 63.2 & 12.8 & {\makecell[l]{it appears to me my new strength and desire .}} & Dickens\\
\bottomrule
\end{tabular}%
\end{small}
\caption{Textual transfer examples}\label{table:ex2}
\end{table*}

\end{document}